\documentclass[runningheads]{llncs}
\usepackage{graphicx}

\usepackage{hyperref}           
\usepackage[normalem]{ulem}     
\useunder{\uline}{\ul}{}        
\usepackage{stfloats}           
\usepackage{subfigure}

\usepackage{cite}
\usepackage{amsmath,amssymb,amsfonts}
\usepackage{algorithmic}
\usepackage{textcomp}
\usepackage{xcolor}

\usepackage[misc]{ifsym}        
\usepackage{appendix}    

\begin{document}
%
\title{CVTGAD: Simplified Transformer with Cross-View Attention for Unsupervised Graph-level Anomaly Detection}
%

%

\author{Jindong Li \and
Qianli Xing \and
Qi Wang (\Letter) \and
Yi Chang}

\toctitle{CVTGAD: Simplified Transformer with Cross-View Attention for UGAD}
\tocauthor{Jindong Li, Qianli Xing, Qi Wang and Yi Chang}

\titlerunning{Published as a conference paper at ECML PKDD 2023}
\authorrunning{Published as a conference paper at ECML PKDD 2023}


\institute{School of Artificial Intelligence, Jilin University, Changchun, 130012, China
\email{jdli21@mails.jlu.edu.cn, \{qianlixing, qiwang, yichang\}@jlu.edu.cn}
}

\maketitle              
\begin{abstract}
Unsupervised graph-level anomaly detection (UGAD) has received remarkable performance in various critical disciplines, such as chemistry analysis and bioinformatics. Existing UGAD paradigms often adopt data augmentation techniques to construct multiple views, and then employ different strategies to obtain representations from different views for jointly conducting UGAD. However, most previous works only considered the relationship between nodes/graphs from a limited receptive field, resulting in some key structure patterns and feature information being neglected. 
In addition, most existing methods consider different views separately in a parallel manner, which is not able to explore the inter-relationship across different views directly. 
Thus, a method with a larger receptive field that can explore the inter-relationship across different views directly is in need.
In this paper, we propose a novel Simplified Transformer with Cross-View Attention for Unsupervised Graph-level Anomaly Detection, namely, CVTGAD. 
To increase the receptive field, we construct a simplified transformer-based module, exploiting the relationship between nodes/graphs from both intra-graph and inter-graph perspectives.
Furthermore, we design a cross-view attention mechanism to directly exploit the view co-occurrence between different views, bridging the inter-view gap at node level and graph level. To the best of our knowledge, this is the first work to apply transformer and cross attention to UGAD, which realizes graph neural network and transformer working collaboratively. Extensive experiments on 15 real-world datasets of 3 fields demonstrate the superiority of CVTGAD on the UGAD task. The code is available at \url{https://github.com/jindongli-Ai/CVTGAD}.


\end{abstract}

\section{Introduction}
Graph data has drawn extensive attention in a variety of domains due to its ubiquity in the real world, such as small molecules, bioinformatics, and social networks \cite{TuDataset}. Graph-level anomaly detection, which is one of the vital research problems in dealing with graph data, aims to identify graphs with anomalous information. 
Usually, anomalous graphs deviate significantly from the normal graphs in the sample \cite{survey_maxiaoxiao}. It has received increasing attention due to its power in various practical applications, such as identifying toxic molecules in chemical compounds analysis, and spotting the molecules with anti-cancer activity in cancer drug discovery \cite{survey_maxiaoxiao,GLocalKD,GLADC,goodd,iGAD}. Despite the remarkable performance achieved by advanced methods and learning paradigms \cite{GLocalKD,GLADC,goodd,iGAD}, there are still some issues that need to be further discussed and addressed in graph-level anomaly detection task. 

\begin{figure}[ht]
\centering
    \includegraphics[width=0.7\textwidth]{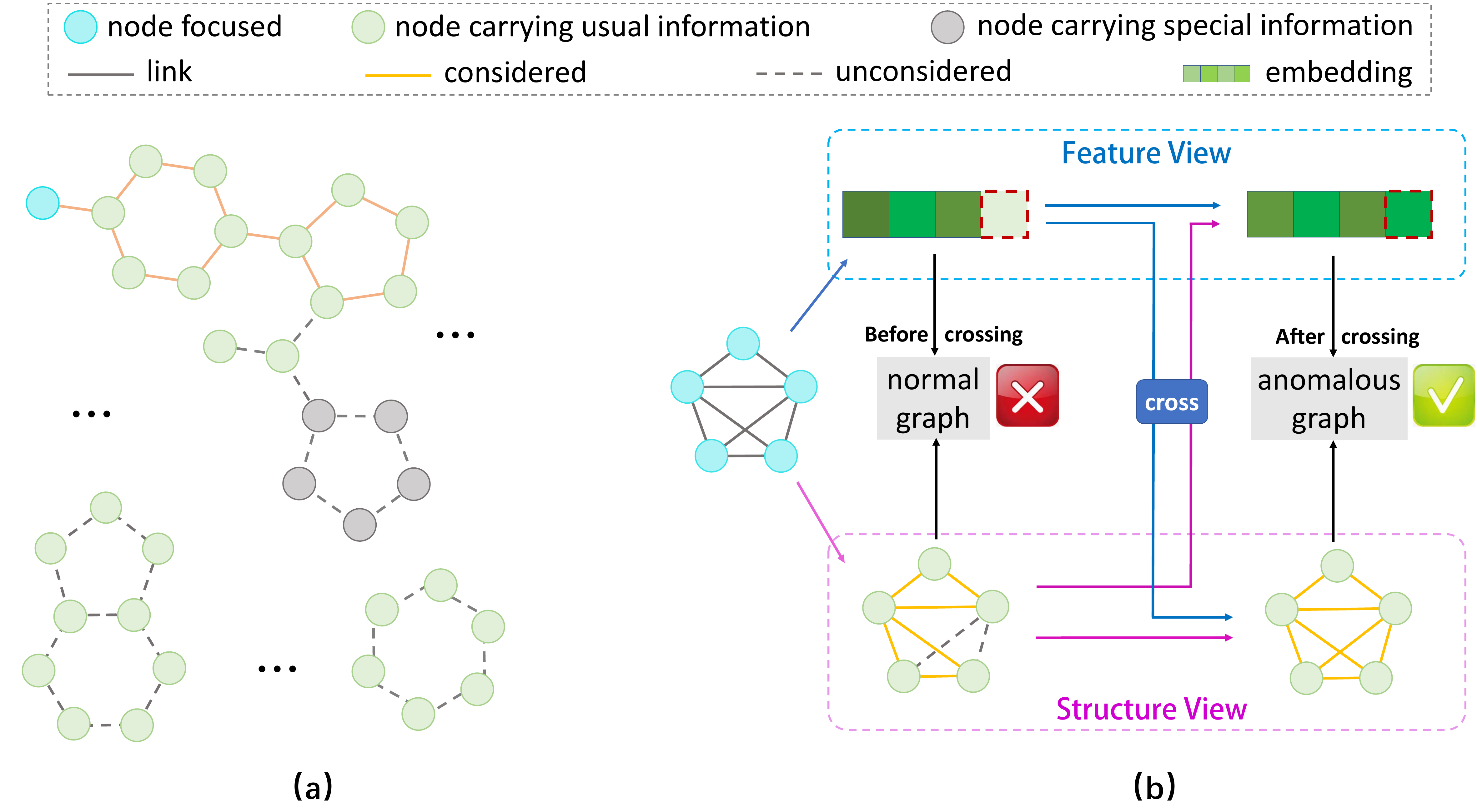}
    \caption{Toy examples to show two major issues: (a) if obtaining the embedding of the current node or graph with a limited receptive field, some nodes or patterns that carry key information would be ignored (e.g., grey in figure); (b) if ignoring the view co-occurrence across different views, some anomalous graphs would not be distinguished accurately.}
    \label{fig1}
\end{figure}

\textbf{Firstly}, existing methods mainly rely on the GNN encoder to obtain the representation of the node/graph \cite{GLocalKD,goodd}. However, due to its limited receptive field, GNNs only consider the local neighbors or sub-graphs (own nodes) of the current node (graph), missing the key anomalous information from the global perspective. For instance, as shown in Figure \ref{fig1} (a), the subgraph composed of grey nodes is difficult to be considered in the existing GNN-based methods with rather limited layers, but this subgraph carries the key information to determine whether the chemical molecular graph is an anomalous graph or not. In addition, the relationship between the current graph and other graphs in the whole dataset (e.g., as shown in the lower part of Figure \ref{fig1} (a)) should also be considered with a larger receptive field. It is because each graph is associated latently and inseparable in the target dataset, which is particularly evident in fields such as chemistry analysis and bioinformatics \cite{GLADC, iGAD}. Existing technologies that simply increase the number of layers for the GNN encoder may cause the over-smoothing problem \cite{2020_AAAI_over_smoothing_problem}. Therefore, it is necessary to design a specific module that increases the considered receptive field of the current node/graph, which can exploit the relationship between nodes/graphs from both intra-graph and inter-graph perspectives, preventing ignoring key information. 

\textbf{Secondly}, some existing methods employ data augmentation techniques to get multiple graph views to enrich the graph information.
The latent mutual agreement between different views is then maximized in the form of a loss item (e.g., InfoNCE-like Loss \cite{goodd}) \cite{contrastive_multi-view_representation_learning_on_graphs__graph_diffusion,GraphCL,Unifying_GCL_with_flexible_contextual_socpes,GCL_with_adaptive_augmentation}. Nevertheless, these works consider different views separately in a parallel manner and simply consider the latent influence between different views via a loss function at the final training step, which is not able to capture the complicated view co-occurrence across different views. In fact, the features of nodes and the structure between nodes influence and entangle each other in the anomalous graph forming process. Specifically, nodes with certain characteristics are more likely to form abnormal links. Similarly, the characteristics of nodes connected by abnormal links may also change accordingly \cite{DOMINANT}. Information from different views reflects different characteristics of an anomalous graph, which are consistent in nature and own view co-occurrence instinctively \cite{2018_multimodal_survey}. For example, as shown in Figure \ref{fig1} (b), it is hard to determine whether the current graph is an anomalous graph or not from any single view alone. Only by comprehensively considering the cross-view information can a more accurate judgment be made. Therefore, there is an urgent need for a novel mechanism to capture such co-occurrence across different views in an explicit way to bridge the inter-view gap of graph-level anomalies.

In this paper, to address the aforementioned issues, we propose a novel Simplified Transformer with Cross-View Attention for Unsupervised Graph-level Anomaly Detection (CVTGAD in short). In concrete, \textbf{for the first issue,} we design a simplified transformer module including projection network, residual network, and transformer to exploit the relationship between nodes/graphs from not only intra-graph but also inter-graph perspectives for increasing the receptive field. \textbf{For the second issue,} we design a cross-view attention mechanism to directly exploit the view co-occurrence between different views (i.e., feature view and structure view), bridging the inter-view gap at node level and graph level. Finally, the anomaly score is obtained by an adaptive anomaly scoring module. \textbf{Our major contributions are summarized as follows:}
\begin{itemize}
    \item We propose a novel simplified transformer framework with cross-view attention for unsupervised graph-level anomaly detection task (CVTGAD). To the best of our knowledge, this is the first work to introduce transformer and cross-attention to unsupervised graph-level anomaly detection, realizing graph neural network and transformer working collaboratively.
    \item We design a simplified transformer with its attention mechanism to capture the relationship between nodes/graphs in both intra-graph and inter-graph perspectives, preventing ignoring key information. In addition, a cross-view attention module is introduced to directly exploit the view co-occurrence across different views, bridging the inter-view gap at both node level and graph level.
    \item We conduct comprehensive experiments against 15 real-world datasets of different fields to demonstrate the effectiveness and superiority of CVTGAD on unsupervised graph-level anomaly detection task.
\end{itemize}

\section{Related Work}

\subsection{Graph-level Anomaly Detection}
Given a graph dataset, graph-level anomaly detection aims to distinguish anomalous graphs from normal graphs \cite{survey_maxiaoxiao}, where the anomalous graphs usually represent very few but essential
patterns. 
Most traditional methods contain two modules: firstly, a graph kernel, such as Weisfeiler-Leman kernel (WL) \cite{WL_graph_kernel} and propagation kernel (PK) \cite{propagation_kernel}, is used to learn node representations. And secondly, an anomaly detector, such as isolation forest (iF) \cite{iF}, one-class support vector machine (OCSVM) \cite{OCSVM}, and local outlier factor (LOF) \cite{2000_LOF}, is applied to detect anomalous graphs based on the acquired graph representations.

In addition, graph neural networks (GNNs) have attracted significant attention due to their remarkable performance in dealing with graph data \cite{representation_learning_on_large_graphs, GCN, how_powerful_are_GNNs__GIN, GAT, goodd}. Thus, various types of GNN are employed as the backbone to conduct graph-level anomaly detection \cite{zhao2021god_OCGIN, GLocalKD, GLADC}. For example, LocalKD \cite{GLocalKD} employs GNN as encoder and achieves random knowledge distillation (KD) \cite{knowledge_distillation_firstpaper, uniformed_students_KD}. The method is achieved by predicting one GNN via training another GNN, where the network weights are all initialized in a random way \cite{GLocalKD}. GOOD-D \cite{goodd} designs a novel graph data augmentation method and employs GIN\cite{how_powerful_are_GNNs__GIN} as encoder to conduct graph-level anomaly detection. However, according to our investigation, graph-level anomaly detection is still under-explored and there are only several research works.

\subsection{Graph Contrastive Learning}
Graph contrastive learning utilizes the mutual information maximization mechanism to obtain a rich representation by maximizing instances with similar semantic information\cite{SSL_Generative_or_contrastive, Graph_SSL_survey}. It has been widely employed for achieving remarkable graph representation learning performance in an unsupervised manner \cite{contrastive_multi-view_representation_learning_on_graphs__graph_diffusion, GCC, InfoGraph, DGI_edge_dropping, GraphCL, Rethinking_and_Scaling_up_GCL, GCL_with_adaptive_augmentation, liu2022beyond, liu2023learning}. For instance, GraphCL \cite{GraphCL} proposes four general data augmentations for graph-structured data to generate pairs for contrastive learning. For graph classification tasks, InfoGraph \cite{InfoGraph} is proposed by maximizing the mutual information between graph-level representations and the substructures-level representations. The substructures-level representations are calculated at different scales.

Some recent works have employed graph contrastive learning to conduct graph-level anomaly detection. For instance, by developing a dual-graph encoder module, GLADC \cite{GLADC} captures node-level and graph-level representations of graphs with  graph contrastive learning techniques. GOOD-D \cite{goodd} detects anomalous graphs based on semantic inconsistencies at different granularities according to the designed hierarchical contrastive learning framework.

\section{Problem Definition}

A graph is denoted as $G = (\mathcal{V}, \mathcal{E})$, where $\mathcal{V}$ is the set of nodes and $\mathcal{E}$ is the set of edges. The topology information of $G$ is represented by an adjacent matrix $\mathbf{A} \in \mathbb{R}^{n \times n}$, where $n$ is the number of nodes. $\mathbf{A}_{i, j} = 1$ if there is an edge between node $v_i$ and node $v_j$, otherwise, $\mathbf{A}_{i, j} = 0$. An attributed graph is denoted as $G = (\mathcal{V}, \mathcal{E}, \mathbf{X})$, where $\mathbf{X} \in \mathbb{R}^{n \times {d_f}}$ represents the feature matrix of node features. Each row of $\mathbf{X}$ represents a node's feature vector with $d_f$ dimensions. The graph set is denoted as $\mathcal{G} = \{G_1, G_2, ..., G_m \}$, where $m$ is the total number of graphs.

In this paper, we focus on the unsupervised graph-level anomaly detection problem. Given a graph set $\mathcal{G}$ containing normal graphs and anomalous graphs, CVTGAD aims to distinguish the anomalous graphs which are different from the normal graphs.

\section{Methodology}
\begin{figure}[!ht]
\centering
\subfigure[The overall framework of CVTGAD.]{
	\label{fig2_a}
	\includegraphics[scale=0.18]{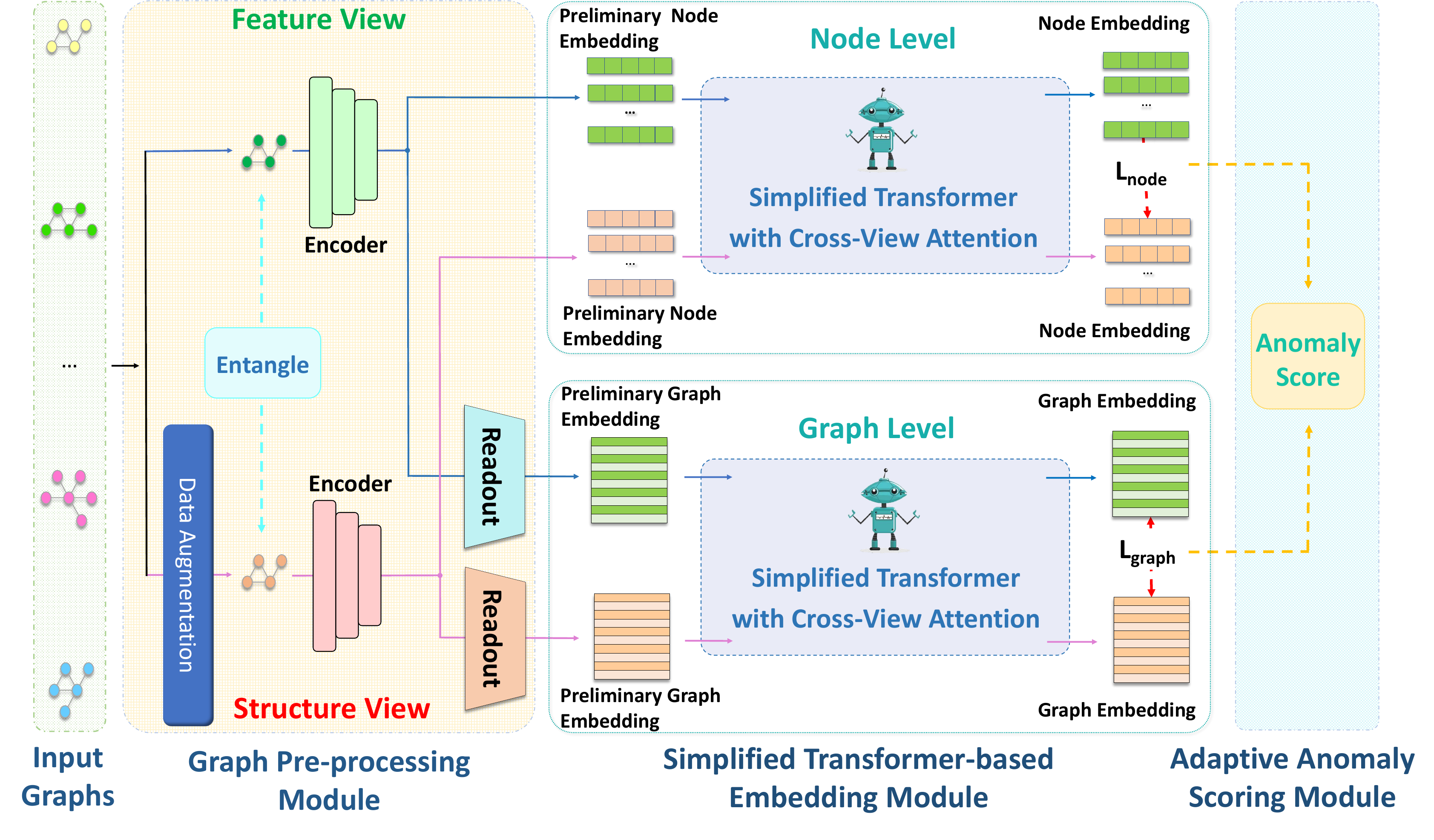}}

\subfigure[The Simplified Transformer with Cross-View Attention.]{
	\label{fig2_b}
	\includegraphics[scale=0.18]{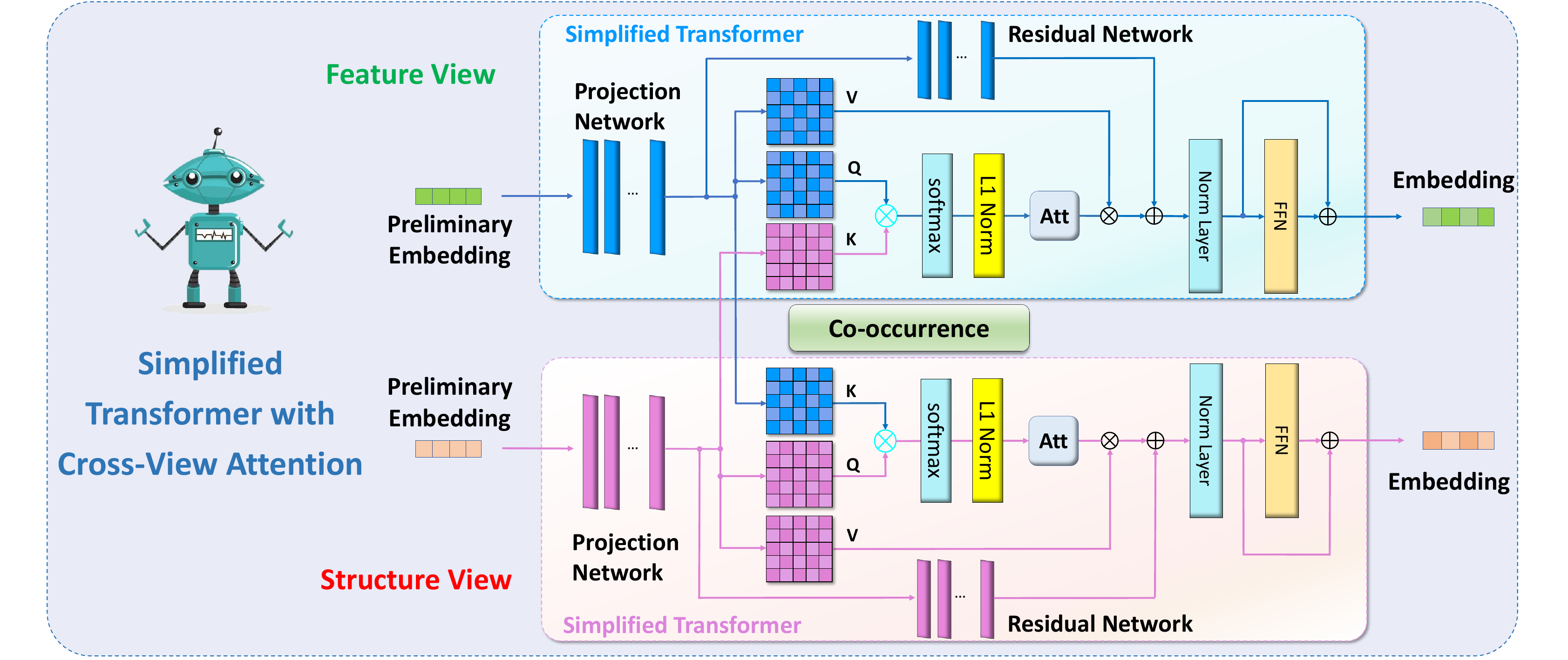}}

\caption{(a) The overview of CVTGAD. The proposed method contains three modules: graph pre-processing module, simplified transformer-based embedding module, and adaptive anomaly scoring module. (b) Specially, we extend the receptive field through a simplified transformer structure and design a cross-view attention mechanism by crossing the matrix $K$.}
\label{fig2}
\end{figure}

In this section, we introduce the proposed method named Simplified Transformer with Cross-View Attention for Unsupervised Graph-level Anomaly Detection (CVTGAD). 
The overall framework of CVTGAD is illustrated in Figure \ref{fig2}, which contains three modules: a graph pre-processing module, a simplified transformer-based embedding module, and an adaptive anomaly scoring module. In the graph pre-processing module, we create two views of each graph by data augmentation. Then, the preliminary node/graph embeddings are calculated by GNN encoders. After that, we exploit the view co-occurrence in the simplified transformer-based embedding module.
In this module, we design a simplified transformer structure with a cross-view attention mechanism to obtain the node/graph embedding with cross-view information.
Finally, an adaptive anomaly scoring module is employed to estimate the anomaly detection score.

\subsection{Graph Pre-processing Module}
In this module, we first generate the feature view and the structure view of each graph. Then, the preliminary node/graph embeddings of two views are calculated, which are used as the input for the simplified transformer-based embedding module. To generate the feature view and structure view of each graph, we adopt the perturbation-free graph augmentation strategy~\cite{goodd, tan2022federated}.

To calculate the preliminary node embedding, we adopt a GNN encoder. Specifically, we employ GIN \cite{how_powerful_are_GNNs__GIN} and GCN \cite{GCN} as GNN encoder in this work. 
Through the GNN encoder, two preliminary node embeddings of feature view and structure view are obtained. As the calculating process of the two kinds of representation is the same, we only show the calculating process of feature view representation obtained from GIN and GCN. 
The propagation rule in the $l$-th layer on the feature view of GIN ($\epsilon = 0$ for simplicity) can be expressed as \cite{how_powerful_are_GNNs__GIN}:
\begin{equation}
    \mathbf{h}^{(f, l)}_{i} = MLP^{(f, l)} \left( \mathbf{h}^{(f, l-1)}_i + \sum_{v_j \in \mathcal{N}(v_i)} \mathbf{h}^{(f, l-1)}_{j} \right),
\end{equation}

where $f$ is the indicator for the feature view. MLP is a multi-layer perception network. $\mathbf{h}^{(f, l-1)}_{i}$ is the preliminary embedding of node $v_i$ in the $l-1$-th layer of feature view, and $\mathcal{N}(v_i)$ is the set of first-order neighbor nodes of node $v_i$.

And the propagation rule in the $l$-th layer on the feature view of GCN can be expressed as \cite{GCN}:
\begin{equation}
    \mathbf{H}^{(f, l)} = \sigma \left(\hat{\mathbf{D}}^{-\frac{1}{2}} \hat{\mathbf{A}} \hat{\mathbf{D}}^{-\frac{1}{2}} \mathbf{H}^{(f, l-1)} \mathbf{W}^{(l-1)} \right),
\end{equation}

where $\hat{\mathbf{A}} = \mathbf{A} + \mathbf{I}_N$ is the adjacency matrix of the input graph $G$ with added self-connections, and $\mathbf{I}_N$ is the identity matrix. $\hat{\mathbf{D}}$ is the degree matrix, $\mathbf{H}^{(f, l-1)}$ is node embedding matrix in the $l-1$-th layer of feature view, $\mathbf{W}^{(l-1)}$ is a layer-specific trainable weight matrix, and $\sigma(\cdot)$ is a non-linear activation function. $\mathbf{h}^{(f, l)}_{i}$ is the $i$-th row of $\mathbf{H}^{(f, l)}$.
And the preliminary node embedding of structure view $\mathbf{h}^{(s, l)}_{i}$ can be calculated in the same way, where $s$ is the indicator for the structure view.

After getting the preliminary embeddings of nodes, a readout function is needed to acquire the preliminary graph embeddings. In this work, we employ global mean pooling as the readout function, which can be represented by:
\begin{equation}
\mathbf{h}^{(f)}_{G} = \frac{1}{|\mathcal{V}_G|} \sum_{v_i \in \mathcal{V}_G} \mathbf{h}^{(f)}_{i}, \quad \mathbf{h}^{(s)}_{G} = \frac{1}{|\mathcal{V}_G|} \sum_{v_i \in \mathcal{V}_G} \mathbf{h}^{(s)}_{i},
\end{equation}

where $\mathcal{V}_G$ is the node set of input graph $G$. $\mathbf{h}^{(f)}_{G}$ and $\mathbf{h}^{(s)}_{G}$ are the feature view and structure view preliminary graph embedding of input graph $G$, respectively. 

\subsection{Simplified Transformer-based Embedding Module}

In this module, we design a novel and simple transformer to adapt to the current graph datasets and unsupervised graph-level anomaly detection task. The proposed simplified transformer architecture comprises a projection network, a residual network, and a transformer (including an attention mechanism, a feed-forward layer, and a norm layer).
After we get the preliminary embeddings of nodes/graphs from different views through graph pre-processing module, we feed them into the projection network, which is achieved by a multi-layer perceptron (MLP) to project them into latent space. The output of it is the input of the residual network and the transformer. The residual network is achieved by a MLP and makes a shortcut between the output of the projection network and the output of the attention layer. 

We achieve feed-forward layer and norm layer with MLP and LayerNorm, respectively\cite{attentionIsAllYouNeed}.
The attention mechanism comprises three parametric matrices: the query matrix $\mathbf{Q} \in \mathbb{R}^{m \times d_k}$, the key matrix $\mathbf{K} \in \mathbb{R}^{m \times d_k}$, and the value matrix $\mathbf{V} \in \mathbb{R}^{m \times d_k}$. 
$m$ is the number of embedding fed into the transformer and $d_k$ is the dimension of embedding. 
For each embedding, the attention matrix $\mathbf{Att} \in \mathbb{R}^{m \times m}$ represents how much it attends to other embeddings, and then transforms the embedding into contextual one \cite{tabtransformer}. $\mathbf{Att}$ is computed as follows:
\begin{equation}
    \mathbf{Att}_{|\mathcal{B}|} = softmax \left( \frac{\mathbf{Q} \mathbf{K}^{T}}{\sqrt{d_k}} \right).
\end{equation}

Each input embedding attends to all other embeddings through the attention mechanism, which could be computed as:
\begin{equation}
    Attention(\mathbf{Q}, \mathbf{K}, \mathbf{V})_{|\mathcal{B}|} = softmax \left( \frac{\mathbf{Q} \mathbf{K}^{T}}{\sqrt{d_k}} \right) \mathbf{V}.
\end{equation}

Note that $\mathcal{B}$ reflects the receptive field. For each node/graph, we calculate the attention of all the nodes/graphs in $\mathcal{B}$. And such a number of nodes/graphs is extremely larger than the existing works. Through this module, the receptive field is extended, which leads to a better representation of nodes/graphs.

Furthermore, we propose a cross-view attention mechanism that aims to directly exploit the view co-occurrence between different views. Specifically, the output of cross-view attention mechanism on the feature view is computed as follows:

\begin{equation}
    Attention(\mathbf{Q}^{(f)}, \mathbf{K}^{(s)}, \mathbf{V}^{(f)})^{(f)}_{|\mathcal{B}|} = softmax-L1\ norm \left( \frac{\mathbf{Q}^{(f)} {\mathbf{K}^{(s)}}^{T}}{\sqrt{d_k}} \right) \mathbf{V}^{(f)},
\end{equation}

where $\mathcal{B}$ is the training/testing batch. Following \cite{SCAN_2018_ECCV}, $softmax-L1\ norm$ means we adopt $softmax$ and $L1\ norm$ to normalize the two dimensions of the attention matrix, respectively. The output of cross-view attention mechanism on the structure view is computed as follows:

\begin{equation}
    Attention(\mathbf{Q}^{(s)}, \mathbf{K}^{(f)}, \mathbf{V}^{(s)})^{(s)}_{|\mathcal{B}|} = softmax-L1\ norm \left( \frac{\mathbf{Q}^{(s)} {\mathbf{K}^{(f)}}^{T}}{\sqrt{d_k}} \right) \mathbf{V}^{(s)}.
\end{equation}

Through the cross-view attention mechanism, we could directly exploit the view co-occurrence between different views,  which bridges the inter-view gap at both node level and graph level.

\subsection{Adaptive Anomaly Scoring Module}
Following \cite{goodd}, we design an adaptive strategy considering both node-level loss and graph-level loss to calculate the anomaly score. 

\textbf{Node-level cross-view contrastive loss.} For an input graph $G$, we construct 
node-level contrastive loss to maximize the agreement between the embedding belonging to different views on the node level:
\begin{equation}
\begin{split}
    \mathcal{L}_{node} = \frac{1}{|\mathcal{B}|} \sum_{G_j \in \mathcal{B}} \frac{1}{2|\mathcal{V}_{G_j}|} \sum_{v_i \in \mathcal{V}_{G_j}} \left[ \emph{l}(\mathbf{h}^{(f)}_{i}, \mathbf{h}^{(s)}_{i}) + \emph{l}(\mathbf{h}^{(s)}_{i}, \mathbf{h}^{(f)}_{i}) \right ] 
\end{split},
\end{equation}

\begin{equation}
    \emph{l}(\mathbf{h}^{(f)}_{i}, \mathbf{h}^{(s)}_{i}) = -log \frac{e^{(sim(\mathbf{h}^{(f)}_{i}, \mathbf{h}^{(s)}_{i})/ \tau)}}{\sum_{v_{k} \in \mathcal{V}_{G_j} \backslash v_i} e^{(sim(\mathbf{h}^{(f)}_{i}, \mathbf{h}^{(s)}_{k}) / \tau)} }.
\end{equation}
In Eq.(8), $\mathcal{B}$ is the training/testing batch and $\mathcal{V}_{G_j}$ is the node set of graph $G_j$. The  calculation of $\emph{l}(\mathbf{h}^{(s)}_{i}, \mathbf{h}^{(f)}_{i})$ and $\emph{l}(\mathbf{h}^{(f)}_{i}, \mathbf{h}^{(s)}_{i})$ are the same, and we show the calculation of $\emph{l}(\mathbf{h}^{(f)}_{i}, \mathbf{h}^{(s)}_{i})$ in Eq.(9) for briefly. In Eq.(9), the $sim(.,.)$ is the function to measure the similarity between different views. In this work, we compute the cosine similarity.

\textbf{Graph-level cross-view contrastive loss.} Similar to node-level loss, we construct a graph-level loss for mutual agreement maximization on graph level:

\begin{equation}
    \mathcal{L}_{graph} = \frac{1}{2|\mathcal{B}|} \sum_{G_i \in \mathcal{B}} \left[\emph{l}(\mathbf{h}^{(f)}_{G_i}, \mathbf{h}^{s}_{G_i}) + \emph{l}(\mathbf{h}^{(s)}_{G_i}, \mathbf{h}^{(f)}_{G_i}) \right],
\end{equation}
\begin{equation}
    \emph{l}(\mathbf{h}^{(f)}_{G_i}, \mathbf{h}^{(s)}_{G_i}) = -log \frac{e^{(sim(\mathbf{h}^{(f)}_{G_i}, \mathbf{h}^{(s)}_{G_i}) / \tau)}}{\sum_{G_j \in \mathcal{B} \backslash G_i} e^{(sim(\mathbf{h}^{(f)}_{G_i}, \mathbf{h}^{(s)}_{G_j}) / \tau)}},
\end{equation}

where notations are similar to node-level loss, and $\emph{l}(\mathbf{h}^{(s)}_{G_i}, \mathbf{h}^{(f)}_{G_i})$ is calculated in the same way as $\emph{l}(\mathbf{h}^{(f)}_{G_i}, \mathbf{h}^{(s)}_{G_i})$.

In the training phase, we employ the adaptive loss function:
\begin{equation}
    \mathcal{L} = \lambda_1 \mathcal{L}_{node} + \lambda_2 \mathcal{L}_{graph},
\end{equation}

where $\lambda_1 = {(\sigma_{node})}^{\alpha}$, and $\lambda_2 = {(\sigma_{graph})}^{\alpha}$. $\sigma_{node}$/$\sigma_{graph}$ is the standard deviations(std) of predicted errors of the node-level/graph-level, where $\alpha \ge 0$ is a hyper-parameter.

In the inference phase, we employ  the normalization method $norm$ to get the final anomaly score:

\begin{equation}
    score_{G_i} = norm({\mathcal{L}_{node}}_{G_i}) + norm( {\mathcal{L}_{graph}}_{G_i}),
\end{equation}

where $norm({\mathcal{L}_{node}}_{G_i}) = ({\mathcal{L}_{node}}_{G_i} - \mu_{node}) / \sigma_{node}$ and $norm({\mathcal{L}_{graph}}_{G_i}) = ({\mathcal{L}_{graph}}_{G_i} - \mu_{graph}) / \sigma_{graph}$. $\mu_{node}$/$\mu_{graph}$ is the mean values of predicted errors of training samples of node-level/graph-level.

\section{Experiment}
In this section, we conduct extensive experiments to validate the effectiveness of our proposed CVTGAD method against 9 baselines on 15 real-world datasets.

\begin{table*}[!ht]
\caption{The statistics of the 15 datasets \cite{TuDataset}.}
\label{datasets}
\centering
\scalebox{0.52}{
    \renewcommand{\arraystretch}{1.6}
    \begin{tabular}{|c|c|c|c|c|c|c|c|c|c|c|c|c|c|c|c|}
    \hline
    \textbf{Dataset}    & \textbf{PROTEINS\_full} & \textbf{ENZYMES} & \textbf{AIDS} & \textbf{DHFR} & \textbf{BZR} & \textbf{COX2} & \textbf{DD} & \textbf{NCI1} & \textbf{IMDB-B} & \textbf{REDDIT-B} & \textbf{COLLAB} & \textbf{HSE} & \textbf{MMP} & \textbf{p53} & \textbf{PPAR-gamma} \\ \hline
    \textbf{Graphs}     & 1113                    & 600              & 2000          & 467           & 405          & 467           & 1178        & 4110          & 1000            & 2000              & 5000            & 8417         & 7558         & 8903         & 8451                \\ \hline
    \textbf{Avg. Nodes} & 39.06                   & 32.63            & 15.69         & 42.43         & 35.75        & 41.22         & 284.32      & 29.87         & 19.77           & 429.63            & 74.49           & 16.89        & 17.62        & 17.92        & 17.38               \\ \hline
    \textbf{Avg. Edges} & 72.82                   & 62.14            & 16.20         & 44.54         & 38.36        & 43.45         & 715.66      & 32.30         & 96.53           & 497.75            & 2457.78         & 17.23        & 17.98        & 18.34        & 17.72               \\ \hline
    \textbf{Node Attr.} & 29                      & 18               & 4             & 3             & 3            & 3             & -           & -             & -               & -                 & -               & -            & -            & -            & -                   \\ \hline
    \end{tabular}%
    }
\end{table*}

\subsection{Experimental Setting}
\textbf{Datasets.} We conduct experiments on 15 public real-world datasets from \cite{TuDataset}, which involved small molecules, bioinformatics, and social networks. Following the setting in \cite{GLocalKD, goodd}, the samples in the minority class or real anomalous class are viewed as anomalies, while the rest are viewed as normal data. Similar to \cite{zhao2021god_OCGIN, GLocalKD, goodd}, only normal data are used for training. The statistics of the datasets are presented in Table \ref{datasets}.

\textbf{Baselines.} To illustrate the effectiveness of our proposed model, we compare CVTGAD with 9 competitive baselines, which can be classified into two groups according to whether contrastive learning is utilized:
(1) For non-contrastive learning-based methods, we select 6 baselines, including PK-OCSVM, PK-iF, WL-OCSVM, WL-iF, OCGIN, and GLocalKD. The PK and WL represent the propagation kernel \cite{propagation_kernel} and the Weisfeiler-Lehman kernel \cite{WL_graph_kernel} separately, which are used to learn the graph embedding. The OCSVM and the iF represent one-class SVM \cite{OCSVM} and isolation forest \cite{iF} separately, which are used as detectors. OCGIN \cite{zhao2021god_OCGIN} and GLocalKD \cite{GLocalKD} are the two latest methods that realize graph anomaly detection in an end-to-end manner;
(2) For contrastive learning-based methods, we select 3 baselines named InfoGraph+iF, GraphCL+iF, and GOOD-D. InfoGraph \cite{InfoGraph} and GraphCL \cite{GraphCL} are two graph embedding methods that use contrastive learning. GOOD-D \cite{goodd} is the latest work that realizes graph anomaly detection in an end-to-end manner using contrastive learning.

\begin{table*}[ht]
\caption{The performance comparison in terms of AUC (in percent, mean value ± standard deviation). The best performance is highlighted in bold, and the second-best performance is underlined. \dag: we report the result from \cite{goodd}.}
\label{overall_performance}
\centering
\scalebox{0.515}{
    \renewcommand{\arraystretch}{1.7}
    \begin{tabular}{ccccccccccc}
    \hline
    Method                 & \textbf{PK-OCSVM\dag} & \textbf{PK-iF\dag} & \textbf{WL-OCSVM\dag} & \textbf{WL-iF\dag} & \textbf{InfoGraph-iF\dag} & \textbf{GraphCL-iF\dag} & \textbf{OCGIN\dag}      & \textbf{GLocalKD\dag}   & \textbf{GOOD-D\dag}     & \textbf{CVTGAD}     \\ \hline
    \textbf{PROTEINS-full} & 50.49±4.92        & 60.70±2.55     & 51.35±4.35        & 61.36±2.54     & 57.47±3.03            & 60.18±2.53          & 70.89±2.44          & \textbf{77.30±5.15} & 71.97±3.86          & {\ul 75.73±2.79}    \\
    \textbf{ENZYMES}       & 53.67±2.66        & 51.30±2.01     & 55.24±2.66        & 51.60±3.81     & 53.80±4.50            & 53.60±4.88          & 58.75±5.98          & 61.39±8.81          & {\ul 63.90±3.69}    & \textbf{67.79±5.43} \\
    \textbf{AIDS}          & 50.79±4.30        & 51.84±2.87     & 50.12±3.43        & 61.13±0.71     & 70.19±5.03            & 79.72±3.98          & 78.16±3.05          & 93.27±4.19          & {\ul 97.28±0.69}    & \textbf{99.39±0.55} \\
    \textbf{DHFR}          & 47.91±3.76        & 52.11±3.96     & 50.24±3.13        & 50.29±2.77     & 52.68±3.21            & 51.10±2.35          & 49.23±3.05          & 56.71±3.57          & {\ul 62.67±3.11}    & \textbf{62.95±3.03} \\
    \textbf{BZR}           & 46.85±5.31        & 55.32±6.18     & 50.56±5.87        & 52.46±3.30     & 63.31±8.52            & 60.24±5.37          & 65.91±1.47          & 69.42±7.78          & {\ul 75.16±5.15}    & \textbf{75.92±7.09} \\
    \textbf{COX2}          & 50.27±7.91        & 50.05±2.06     & 49.86±7.43        & 50.27±0.34     & 53.36±8.86            & 52.01±3.17          & 53.58±5.05          & 59.37±12.67         & {\ul 62.65±8.14}    & \textbf{64.11±3.22} \\
    \textbf{DD}            & 48.30±3.98        & 71.32±2.41     & 47.99±4.09        & 70.31±1.09     & 55.80±1.77            & 59.32±3.92          & 72.27±1.83          & \textbf{80.12±5.24} & 73.25±3.19          & {\ul 77.82±1.60}    \\
    \textbf{NCI1}          & 49.90±1.18        & 50.58±1.38     & 50.63±1.22        & 50.74±1.70     & 50.10±0.87            & 49.88±0.53          & \textbf{71.98±1.21} & 68.48±2.39    & 61.12±2.21          & {\ul 69.07±1.15}          \\
    \textbf{IMDB-B}        & 50.75±3.10        & 50.80±3.17     & 54.08±5.19        & 50.20±0.40     & 56.50±3.58            & 56.50±4.90          & 60.19±8.90          & 52.09±3.41          & {\ul 65.88±0.75}    & \textbf{70.97±1.35} \\
    \textbf{REDDIT-B}      & 45.68±2.24        & 46.72±3.42     & 49.31±2.33        & 48.26±0.32     & 68.50±5.56            & 71.80±4.38          & 75.93±8.65          & 77.85±2.62          & \textbf{88.67±1.24} & {\ul 84.97±2.41}    \\
    \textbf{COLLAB}        & 49.59±2.24        & 50.49±1.72     & 52.60±2.56        & 50.69±0.32     & 46.27±0.73            & 47.61±1.29          & 60.70±2.97          & 52.94±0.85          & {\ul 72.08±0.90}    & \textbf{72.92±1.44} \\
    \textbf{HSE}           & 57.02±8.42        & 56.87±10.51    & 62.72±10.13       & 53.02±5.12     & 53.56±3.98            & 51.18±2.71          & 64.84±4.70          & 59.48±1.44          & {\ul 69.65±2.14}    & \textbf{70.30±2.90} \\
    \textbf{MMP}           & 46.65±6.31        & 50.06±3.73     & 55.24±3.26        & 52.68±3.34     & 54.59±2.01            & 54.54±1.86          & \textbf{71.23±0.16} & 67.84±0.59          & 70.51±1.56          & {\ul 70.96±1.01}    \\
    \textbf{p53}           & 46.74±4.88        & 50.69±2.02     & 54.59±4.46        & 50.85±2.16     & 52.66±1.95            & 53.29±2.32          & 58.50±0.37          & {\ul 64.20±0.81}    & 62.99±1.55          & \textbf{67.58±3.31} \\
    \textbf{PPAR-gamma}    & 53.94±6.94        & 45.51±2.58     & 57.91±6.13        & 49.60±0.22     & 51.40±2.53            & 50.30±1.56          & \textbf{71.19±4.28} & 64.59±0.67          & 67.34±1.71          & {\ul 68.25±4.66}    \\ \hline
    Avg.Rank               & 8.73              & 7.73           & 6.93              & 7.47           & 6.53                  & 6.93                & 3.60                & 3.27                & 2.40                & 1.40                \\ \hline
    \end{tabular}
}
\end{table*}

\textbf{Evaluation Metrics.} We evaluate methods using a popular graph-level anomaly detection metric, i.e., the area under the receiver operating characteristic curve (AUC) following \cite{GLocalKD, goodd, GLADC}. Higher AUC values indicate better anomaly detection performance. 

\textbf{Implementation Details.} In practice, we implement CVTGAD with Pytorch\renewcommand{\thefootnote}{1}\footnote{\noindent https://pytorch.org/}. In order to reduce the uncertainty of this process and ensure reproducibility, we set random seeds explicitly as much as possible following \cite{goodd}. We achieve the projection networks and residual networks with two-layer MLP. 

\subsection{Overall Performance Comparison}
The overall performance of all methods w.r.t AUC against 15 datasets is shown in Table \ref{overall_performance}. As shown in Table \ref{overall_performance}, our proposed CVTGAD outperforms all baselines on 9 datasets and achieves the second-best performance on 6 datasets. And CVTGAD achieves the first place in average rank among all comparative methods against 15 datasets as shown in the last row in Table \ref{overall_performance}. The graph kernel-based methods achieve the worst performance. It may be because they fail to capture regular patterns and key information. The GCL-based methods achieve a modest performance, indicating that GCL-based methods are competitive on this task. These results demonstrate the superiority and effectiveness of CVTGAD on graph-level anomaly detection in different
fields.

\begin{figure*}[!ht]
\centering
    \includegraphics[width=0.7\textwidth]{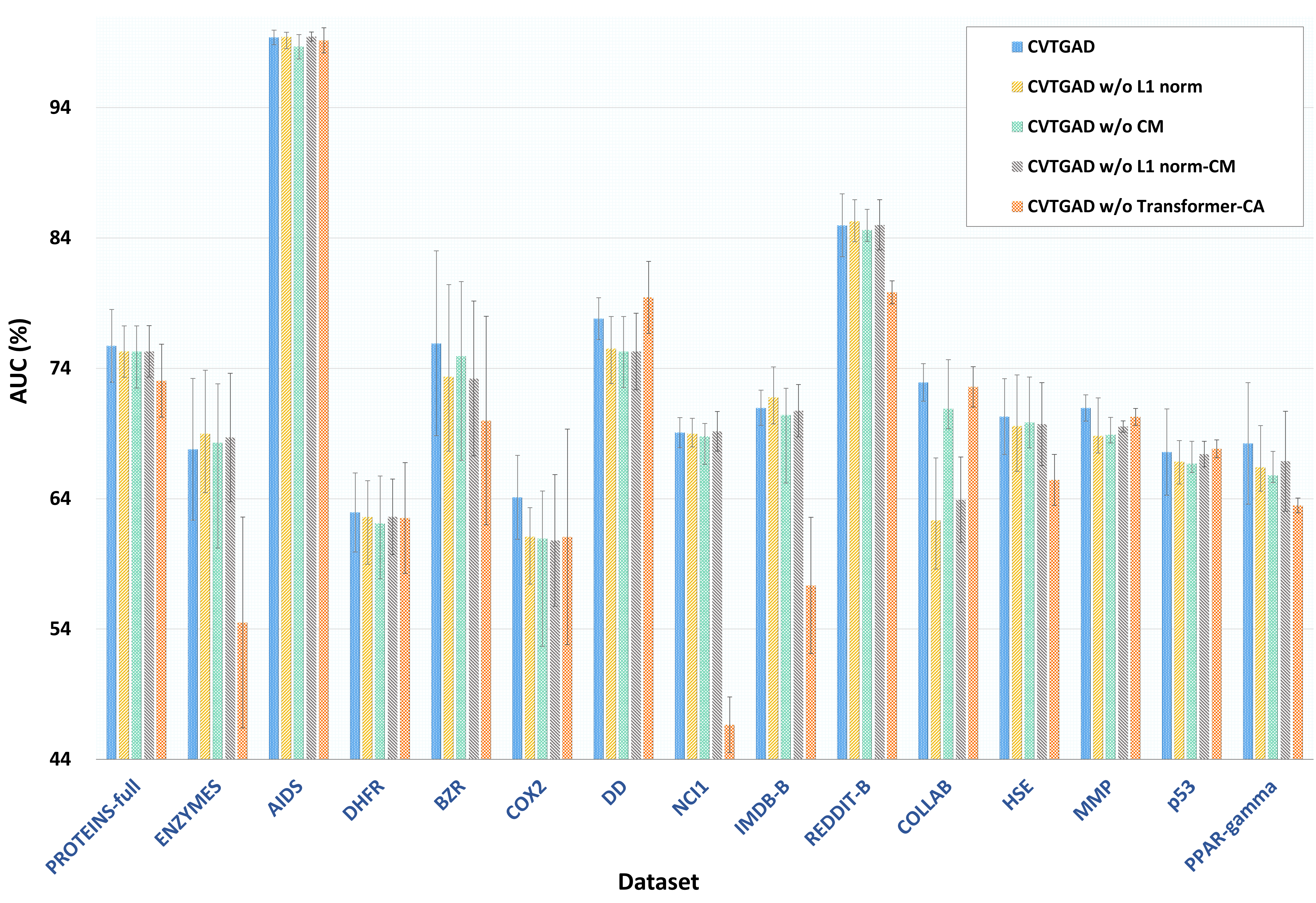} 
    \caption{The comparison of ablating different key components in terms of AUC.}
    \label{ablation_fig}
\end{figure*}

\begin{table}[ht]
\caption{Ablation study results of key components in terms of average rank.}
\label{ablation_tab}
\centering
\scalebox{0.6}{
    \renewcommand{\arraystretch}{1.7}
    \setlength{\tabcolsep}{1mm}
    \begin{tabular}{|c|c|c|c|l|l|}
    \hline
             & CVTGAD & CVTGAD w/o L1 norm & CVTGAD w/o CM & CVTGAD w/o L1 norm-CM & CVTGAD w/o Transformer-CA \\ \hline
    Avg.Rank & 1.40   & 1.87               & 1.80                       & \multicolumn{1}{c|}{1.80}          & \multicolumn{1}{c|}{2.80} \\ \hline
    \end{tabular}
}
\end{table}

\subsection{Ablation Study--Effects of Key Components}

To get a better understanding of the proposed model CVTGAD, we conduct ablation study experiments on 15 datasets to investigate the effects of the three key components: simplified transformer, crossing matrix, and L1 norm. For convenience, let CVTGAD w/o L1 norm, CVTGAD w/o CM, CVTGAD w/o L1 norm-CM, CVTGAD w/o Transformer-CA denote the customized variants of CVTGAD without L1 norm, crossing matrix operation, L1 norm and crossing matrix operation, and simplified transformer module with cross-view attention module, respectively. The experimental results are illustrated in Figure \ref{ablation_fig} and Table \ref{ablation_tab}. We can observe that CVTGAD consistently achieves the best performance against other variants, demonstrating that the simplified transformer with cross-view attention is necessary to yield the best detection results. Compared with CVTGAD, the poor performance of CVTGAD w/o CM proves the significance of directly exploiting view co-occurrence between different views. And the poor performance of CVTGAD w/o Transformer-CA proves the importance of directly exploiting the relationship between nodes/graphs in both intra-graph and inter-graph perspectives. By observing CVTGAD w/o L1 norm, CVTGAD w/o CM, and CVTGAD w/o L1 norm-CM simultaneously, we can draw a preliminary conclusion: L1 norm and crossing operation are both important for implementing the cross-view attention mechanism. L1 norm has little significance for the self-attention mechanism and  is even harmful to it.

\subsection{Hyper Parameter Analysis and Visualization}

\begin{figure*}[!ht]
\centering
    \includegraphics[width=0.65\textwidth]{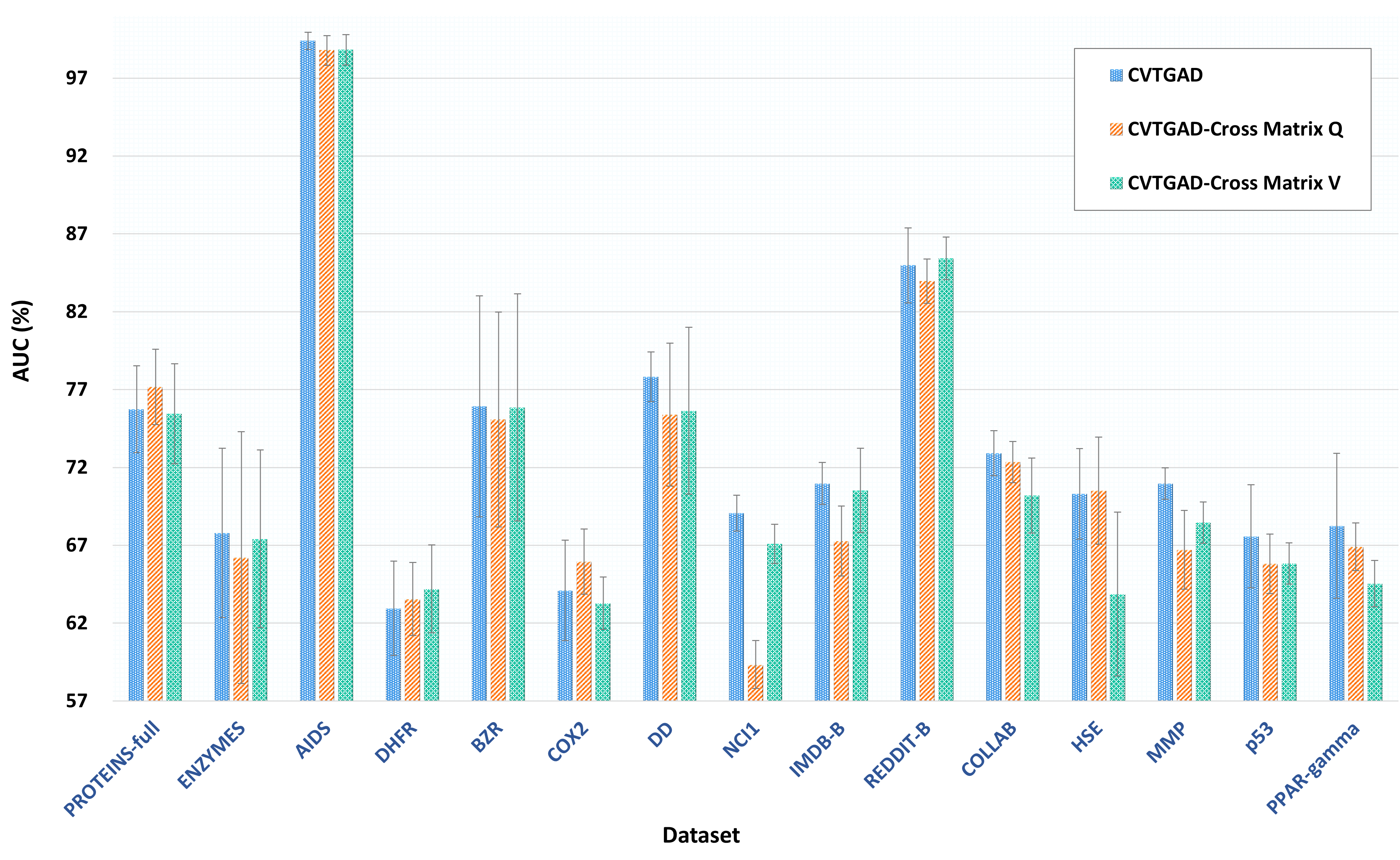} 
    \caption{The comparison of crossing different matrices in cross-view attention mechanism in terms of AUC.}
    \label{Cross-KQV_fig}
\end{figure*}

\begin{table}[ht]
\caption{The comparison of crossing different matrices in cross-view attention mechanism in terms of average rank.}
\label{Cross_KQV_tab}
\centering
\scalebox{0.6}{
    \renewcommand{\arraystretch}{1.7}
    \setlength{\tabcolsep}{5mm}
    \begin{tabular}{|c|c|c|c|}
    \hline
    Model    & CVTGAD & CVTGAD-Cross Matrix Q & CVTGAD-Cross Matrix V \\ \hline
    Avg.Rank & 1.40   & 1.80                  & 1.87                  \\ \hline
    \end{tabular}
}
\end{table}

\subsubsection{The Effect of Different Matrices in Cross-View Attention.}
In cross-view attention, there are three parametric matrices that can cross. We conduct experiments on 15 datasets to investigate the impact of crossing different matrices on detection performance. For convenience, let CVTGAD-Cross Matrix $\mathbf{Q}$ and CVTGAD-Cross Matrix $\mathbf{V}$ denote the customized variants of CVTGAD by crossing matrix $\mathbf{Q}$ and crossing matrix $\mathbf{V}$ in cross-view attention, respectively. The experimental results are illustrated in Figure \ref{Cross-KQV_fig} and Table \ref{Cross_KQV_tab}. We find that: 1) the performance of crossing matrix $\mathbf{K}$ is better than the other two operations; 2) the performance of crossing matrix $\mathbf{K}$ on attribute graphs is more pronounced than on plain graphs. We think it may be that attribute graphs tend to rely on explicit information, while plain graphs can only rely on implicit information, and crossing operation is better at capturing implicit information.

\begin{figure*}[!ht]
\centering
    \includegraphics[width=0.7\textwidth]{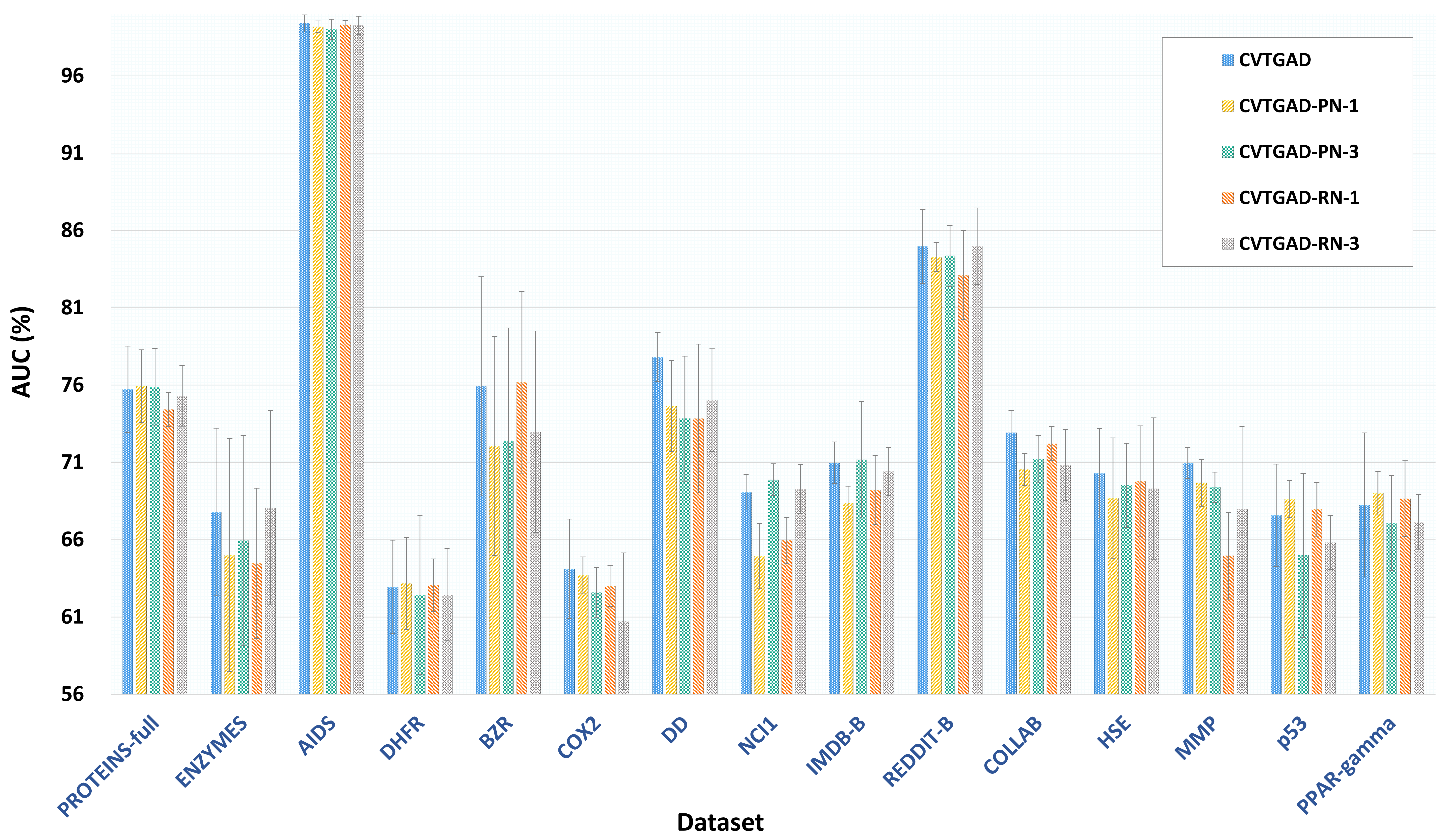} 
    \caption{The comparison of different numbers of layers for projection network and residual network in terms of AUC.}
    \label{hyper_parameter_fig}
\end{figure*}

\begin{table}[ht]
\caption{performance of different numbers of layers for the projection network and residual network in terms of average rank.}
\label{hyper_parameter_tab}
\centering
\scalebox{0.6}{
    \renewcommand{\arraystretch}{1.7}
    \setlength{\tabcolsep}{5mm}
    \begin{tabular}{|c|c|c|c|c|c|}
    \hline
    Model    & CVTGAD & CVTGAD-PN-1 & CVTGAD-PN-3 & CVTGAD-RN-1 & CVTGAD-RN-3 \\ \hline
    Avg.Rank & 1.40   & 1.73        & 1.87        & 1.60        & 1.87        \\ \hline
    \end{tabular}
}
\end{table}

\subsubsection{Number of Layers for Projection Network and Residual Network.} 
We conduct experiments on 15 datasets to investigate the impact of layer number on the projection network and residual network. For convenience, let CVTGAD-PN-1 and CVTGAD-PN-3 denote the customized variants of CVTGAD by achieving the projection network with 1 layer- and 3 layers- MLP, respectively. Let CVTGAD-RN-1, and CVTGAD-RN-3 denote the customized variants of CVTGAD by achieving the residual network with 1 layer- and 3 layers- MLP, respectively. The experimental results are illustrated in Figure \ref{hyper_parameter_fig} and Table \ref{hyper_parameter_tab}. Based on the performance of different variants of CVTGAD, we can conclude that setting the number of layers for the projection network and residual network to 2 is the best choice. 
We think that the poor performance obtained when the number of layers is 1 is due to insufficient expression ability of the network, and the poor performance when the number of layers is 3 is due to overfitting caused by too deep layers.

\textbf{Visualization}. We use t-SNE\cite{tSNE_paper} to visualize the embeddings learned by CVTGAD. We can observe that it is difficult to directly distinguish anomalous graphs from normal graphs by relying solely on feature space or structure space. But there is a clear scoring boundary (anomaly score=18), which results in a good performance on anomaly detection. This shows the effectiveness of CVTGAD.

\begin{figure}[!ht]
\centering
\subfigure[Graph-level f-view.]{
	\label{fig6_a}
	\includegraphics[scale=0.24]{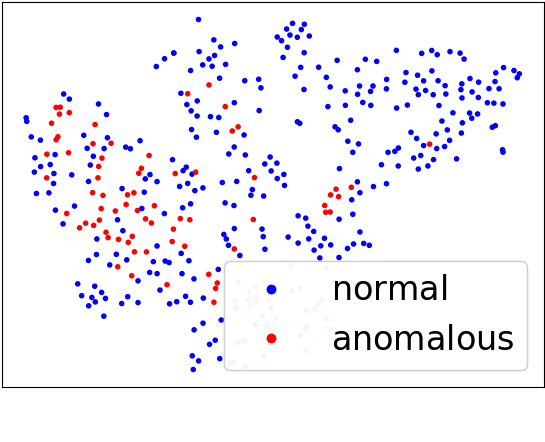}}
\quad
\subfigure[Graph-level s-view.]{
	\label{fig6_b}
	\includegraphics[scale=0.24]{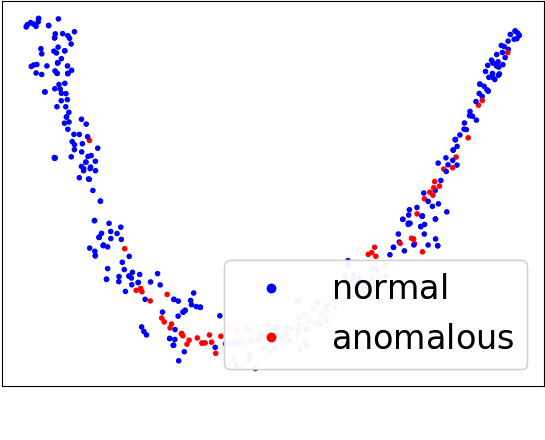}}
\subfigure[Anomaly score.]{
	\label{fig6_c}
	\includegraphics[scale=0.25]{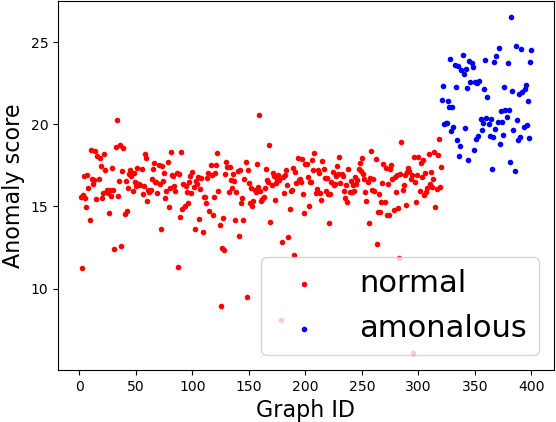}}
\caption{Visualization on AIDS dataset (a) and (b): visualization of testing sample embeddings of feature view (f-view) and structure view (s-view) on graph-level by t-SNE. (c): anomaly score on testing samples.}
\label{fig6}
\end{figure}

\section{Conclusion}
In this paper, we propose a novel framework named CVTGAD, which mainly employs a simplified transformer with a proper receptive field to capture key information and employs a cross-view attention mechanism to directly exploit the view co-occurrence across different views. To the best of our knowledge, we are the first to introduce a transformer and cross attention to the UGAD task, achieving graph neural network and transformer working collaboratively. Extensive experiments demonstrate the superiority of CVTGAD against 15 real-world datasets of different fields.

\section{Acknowledgements}
This work is supported by the Youth Fund of the National Natural Science Foundation of China (No. 62206107).

\newpage
\section*{Ethical Statement}
The following statement outlines the ethical considerations that were taken into account during the research process.

\textbf{Data Collection.} 
In the experimental part, we used a publicly available dataset. And the public dataset has preprocessed the information involved in the data, so there are no issues of confidentiality and privacy.

\textbf{Protection of Participants.}
Throughout the research process, there were no additional participants except the authors. The personal information of all personnel is not related to the experiment. The experiment only used information from public dataset.

\textbf{Data Analysis.} 
Our data analysis is only from the perspective of algorithmic metrics, without any discrimination or illegal tendencies.

\textbf{Conflict of Interest.} 
We declare that they have no conflicts of interest that may have influenced the research.

\textbf{Research Involving Animals.}
This study does not involve the use of animals.

\textbf{Cultural Sensitivity.}
The research team was aware of the potential cultural biases that could have an impact on the study results. To ensure cultural sensitivity, the research team worked with participants from diverse cultural backgrounds and used culturally appropriate language in the consent form and data collection procedures.

\textbf{Beneficence.}
The research team considered the potential benefits and harms of the study. The research team made efforts to minimize any potential harms to participants while maximizing the potential benefits to both individuals and society.

\bibliographystyle{splncs04}
\newpage
\bibliography{CVTGAD.bib}

\newpage

\end{document}